# Toward Game Level Generation from Gameplay Videos

Matthew Guzdial, Mark O. Riedl
School of Interactive Computing
Georgia Institute of Technology
{mguzdial3; riedl}@gatech.edu

## ABSTRACT
Algorithms that generate computer game content require game design knowledge. We present an approach to automatically learn game design knowledge for level design from gameplay videos. We further demonstrate how the acquired design knowledge can be used to generate sections of game levels. Our approach involves parsing video of people playing a game to detect the appearance of patterns of sprites and utilizing machine learning to build a probabilistic model of sprite placement. We show how rich game design information can be automatically parsed from gameplay videos and represented as a set of generative probabilistic models. We use Super Mario Bros. as a proof of concept. We evaluate our approach on a measure of playability and stylistic similarity to the original levels as represented in the gameplay videos.

## Categories and Subject Descriptors
I.2.1 [Artificial Intelligence]: Applications and Expert Systems—Games

## 1. INTRODUCTION
Procedural content generation (PCG) has been used to automatically create levels, maps, weapons, background scenery, and music for computer games [5, 12]. Intelligent generative methods must be provided with high quality design knowledge to create compelling content. Often this design knowledge is provided in the form of hand-coded heuristics [4] or evaluation functions [12]. Hand-coded heuristics and evaluation functions provide a PCG system with intuition about what makes a particular type of content *good* but also biases the system toward the particular beliefs of the system designer. Alternatively, design knowledge can be extracted from the game itself. For example, a system can parse game level files to extract level design patterns [4, 10]. However, such approaches make use of hand-authored information to both parse games' unique file structures and to ascribe meaning to the collected structures.

In this work we propose an alternative source of design knowledge: gameplay videos. Acquiring design knowledge from gameplay videos has a number of advantages. (1) Gameplay videos exist within a number of set formats that are largely interchangeable, meaning that an algorithm does not need to be rewritten to handle new asset formats. (2) Gameplay videos include a player's reaction to game assets, meaning that such systems can learn not only design information but also its effect on player experience. (3) With the advent of "Let's Plays" and "Long Plays," in which individuals make video recordings of their game playthroughs publicly available, a large corpora of gameplay video data exists for many different games.

We present an approach to acquire game level design knowledge from gameplay videos of Super Mario Bros. While we focus on this well understood game for our preliminary exploration, the technique we present can extend to other two dimensional platformer games. By applying the technique across a number of different platformer games, a system can theoretically learn *genre* knowledge, which can be beneficial for procedurally creating novel games of a given genre. Our technique may also extend to other game genres beyond platformers.

As proof of concept, we focus on learning design knowledge from Super Mario Bros. in isolation. We focus on two specific aspects of learning level design knowledge from video data: (1) determining what to learn about level layout, and (2) a representation of level design knowledge in a reusable form that affords generation. To process gameplay videos, we use OpenCV [9], a freely-available, open source computer vision toolkit, to process each frame of each video.

For the first problem, determining what to learn, we present a technique to identify and categorize *high interaction areas* in a game level as a means of showcasing the affordance of user interactions captured in gameplay video. A high interaction area is a section of a level in which players spend significantly more time than in other sections. This may be because the area is more visually interesting, more rewarding (e.g., a lot of coins or power ups), more challenging (e.g., a jumping puzzle), or requires more navigation to traverse. We extract sequences of high interaction from the full video trace and use OpenCV to extract features from these sequences by parsing the placement of sprites.

For the second problem, representation of design knowledge, we present a technique for learning generative probabilistic models of level sections. A generative probabilistic model represents a section of game level as a set of distributions over sprites, their positions, and relationships with one another. Frames of video that cover sections identified as high interaction areas are clustered together to provide a training set for each graphical probabilistic model. In addition to representing knowledge about how high interaction areas are laid out, these models also act like templates, allowing new areas to be generated. Our technique extends the work of Kalogerakis et al [6], which was originally developed to procedurally generate variations of 3D graphical models. We show how the probabilistic models learned by our system can be used to generate new high interaction areas through a recursive walk of the model.

To the best of our knowledge this work represents the first attempt to automatically learn game design knowledge from gameplay video data. The contributions of our work are as follows: (1) an approach to automatically identify and categorize level sections to be modeled from gameplay video, (2) an approach to automatically generate new level sections from these models, and (3) an initial evaluation of our generated level sections on playability and stylistic metrics.

## 2. RELATED WORK

PCG systems take in design knowledge, utilize that design knowledge to create assets or mechanics, and output game content. Approaches include evolutionary search, rule-based systems and instantiating content from probability tables [12].

Automated game design knowledge acquisition is a process of deriving understanding of some facet of a computer game for the purposes of analysis or procedural generation. Dahlskog and Togelius [3]; and Snodgrass and Ontañón [10] both make use of levels directly from the original Super Mario Bros. game. The former generates new levels via evolutionary computation with the original levels as part of the scoring criteria. The latter make use of hierarchical levels of Markov chains, representing both low and high level patterns in the original levels. Both approaches have been shown to produce good results, but require that the source code for levels be available. In addition, both approaches involve hand defined heuristics or patterns. Our technique avoids the potential pitfalls of too little human design knowledge, while not requiring additional authoring, and is theoretically extendible to a wide range of game genres.

Machine vision is not often applied to computer games. Mnih et al. [7] used deep convolutional networks to learn how to the play Atari games from pixel input. Although this system and our own work both use machine vision to process pixels over time, our system focuses on extracting design principles instead of merely learning how to play the game.

The generative probabilistic model we use to represent game level design knowledge and for use in generating new level sections is inspired by work done in the field of computer graphics. Kalogerakis et al. [6] describes work on learning probabilistic grammars from which hundreds of new 3D graphical models can be produced from a small training set. Their technique was semi-automated in the sense that they made use of human tagging of parts of the input 3D model ("arm", "head", "torso", etc.). We contribute a fully automated extension of their approach, which has been customized to discrete sprite-based game worlds.

## 3. LEARNING DESIGN KNOWLEDGE

In this section we present an approach to learning and modeling game level design knowledge from gameplay video. Beyond patterns of assets, gameplay video demonstrates how these patterns of assets impact player behavior. We highlight the potential of using gameplay video as a source of design knowledge by automatically extracting areas of Super Mario Bros. in which the player spends above average time. We learn a generative probabilistic model of these high interaction areas, which acts like a template, allowing us to generate new level sections that would be anticipated to also take the player longer to traverse. This evidence-based approach is significant in that it only relies on how player behavior observably changes, which may be different from the original game designer's intentions.

Our specific technique for using gameplay video to acquire game level design knowledge consists of the following steps. First, we collect gameplay video data from online video portals. Using the OpenCV [9] machine vision toolkit, we scan each frame of each video, recording the sprites and their locations. Our system scans the records for periods in which players are significantly slower in progressing through the level as identified by periods in which the difference in frame features is small. We assume that high interaction areas have special significance from a level design perspective and target these areas specifically to learn game level design knowledge.

Having identified sections of game levels of interest—in this case areas of increased player interaction—our system proceeds to create probabilistic models of level design rules for the identified clusters of frames. The result is game level design knowledge represented as a probabilistic grammar that expresses the likely arrangements of level assets (sprites) on the screen. Following Kalogerakis et al. [6], we search each frame for identical and adjacent sprites to build shapes of tiled sprites and then search for probabilistic relationships between these sprite shapes.

The probabilistic models used to model different sections of game levels can be used to generate new game level sections. We use the probabilistic grammar and the original examples to search through all possible configurations of tiles for those that meet a user-supplied value of playability and approximate similarity of style to the original level sections.

We describe these three sequential processes in detail in the sections below.

### 3.1 Level Section Categorization

The purpose of the first step of our approach is to pull usable level design information from the gameplay video data. In particular our process identifies the distinct sections of level design that are likely to have interesting design elements to learn. Toward future work to generate whole levels from level sections we use the notion of *interaction value*—the time a player spends in a particular section of a game level—as the metric for interestingness and track its variance across level sections. We collected eleven gameplay videos as an initial dataset, all of which demonstrated individuals playing through the entire game.

Machine vision has progressed to the point where open toolkits such as OpenCV provide sufficient results for straightforward vision tasks. Classic computer games are amenable to machine vision due to the regularity and consistency of appearance of objects on the screen. In games this regularity and consistency is referred to as *tiling* and the individual objects on screen referred to as *sprites*. The implication of this tiling is that we can consistently identify everything in gameplay video of Super Mario Bros. with a database of 102 sprites. While this represents a significant task in terms of compiling these sprites, it only needs to occur once to parse any amount of footage. Parsing each gameplay video with OpenCV outputs a description of each frame as a list of sprites and their xy coordinates, which our process uses as its representation of a level section.

The process next segments the frame data per video into distinct sections of level. This is accomplished by identifying edges of level sections where the contents of a frame differ by more than 10% of the total possible differences. This technique identifies sections of level design and determines the start and end point of levels as moments where one frame differs completely from the last. Under this definition our system also identifies deaths of Mario as level endpoints because the screen turns black. To avoid this issue we make use of only expert gameplay data of Super Mario Bros., in which the player makes it through all levels without dying.

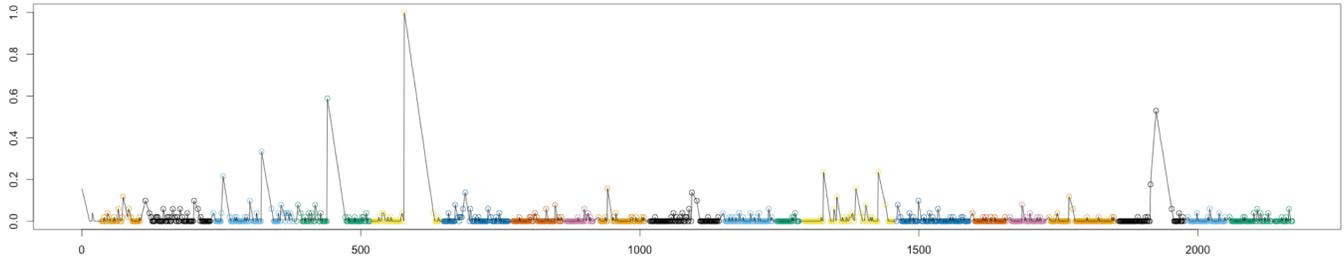

**Figure 1**. A graph of a video's level sections vs intensity values (units of 25 seconds), with gameplay levels coded in different colors.

The majority of games have some variance of qualities across game levels, and Super Mario Bros. is no exception. We measure player interaction time over a level section as a means of tracking this variance. These values of player interaction are measured by number of frames the player spends on each of the level sections identified in the previous step. Figure 1 is a graph of these interaction values over an entire gameplay video where each point on the x-axis a distinct level section within a level and the values along the y-axis are the interaction values of those sections.

We next focus on categorization of our defined level sections. Categorization is used to identify level sections that share design patterns. At this time, we only categorize high interaction sections. We identify these sections as those whose interaction times are above average for an individual player. These sections are of interest as each player spends the largest proportion of their playtime in these sections. They therefore serve as an appropriate focus for initial work with our novel generation approach.

Across high interaction areas, we make the assumption that different sprites and different numbers of these sprites indicate that different design rules are at play. For example, we are more likely to see higher numbers of "block" sprites in underground levels than in typical levels. Our system clusters high interaction level sections using K Means++ wherein K is estimated with the distortion ratio (c.f., [1, 8]). For a distance measure we used the Euclidean metric across vectors of count data for each sprite. To avoid issues related to variance in level section size, we only made use of the data from the first frame of each section.

Our approach finds 21 clusters of frames within the eleven full playthroughs of Super Mario Bros. we collected. These clusters vary from 2 to 250 frames in size with a median value of 35 frames. From a human perspective we find these clusters to be largely cohesive. 18 of the clusters have a clear theme that can be recognized by visual inspection. Examples of themes include "underwater," "underground," or "treetops." Only three clusters do not have a clear theme to them and are also the three largest clusters. We find these clusters to be sufficient for this approach as any disparity of theme within the clusters can be identified via further sub-clustering during probabilistic modeling as discussed in the next section.

### 3.2 Probabilistic Model Construction

Design knowledge learned from the clusters in the previous section is represented in the form of a generative probabilistic graphical model that relates groupings of sprites to each other for distinct high interaction areas. The probabilistic model can then be used to generate brand new level sections that are implied by the model. Our model is inspired by Kalogerakis et al's work on generating 3D models from an initial sample set. Their approach constructs a probabilistic graphical model that determines the values of latent or hidden variables that represent design rules (e.g., pipes go on top of ground or 3D models with arms have torsos) from observed variables. We use the same underlying intuitions, but due to key differences in domain make a series of alterations to our model, how we learn the model, and how we use it to generate new level sections. The key differences due to domain are (a) a lower reliance on human authored information and (b) its two dimensional, sprite-based nature. Figure 2 is a visual representation of the tree-like structure of our final model and Table 1 identifies its major components and their human interpretation. Our model uses three types of nodes, (G, D, N), to collect observable information from an initial data set, each of which collects unique information. In addition there are two nodes, (S, L), that represent latent or hidden information within the initial dataset and are derived from the observed information.

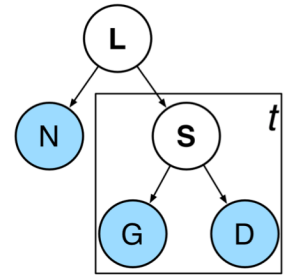

**Figure 2.** A visualization of the probabilistic model. Blue nodes are those derived from observable variables.

**Table 1.** Brief descriptions of each node by their interpretation.

| Node | Interpretation |
|---|---|
| L | Level design style (latent) |
| S | Shape style per sprite $t$ (latent) |
| N | Count data for the different categories of sprites found in an initial level section |
| G | Geometric information for patterns of sprite $t$ |
| D | Relational information between patterns of sprite $t$ and all other $t$ values. |

The G node collects the geometric information of a shape composed of sprites of type $t$. Figure 3 (top) shows an example of the rectangular shape composed of the "bark" sprite that serves as the basis for a G Node highlighted in the center of the image. However, it is important to note that Figure 3 (top) actually contains three detected "bark" shapes, each of which is the basis for distinct G Nodes. Each shape is derived by joining sprites of the same type according to adjacency such that there are no disjoint sprites. By iterating through each level section within a high interaction category and across each sprite the model collects a set of G nodes for each sprite type $t$. This set may contain duplicates in terms of the same shape, which we will generalize over in a later step. Since our domain's atomic parts are sprites our G node makes use of the categorical information of its sprite type and continuous position information.[1]

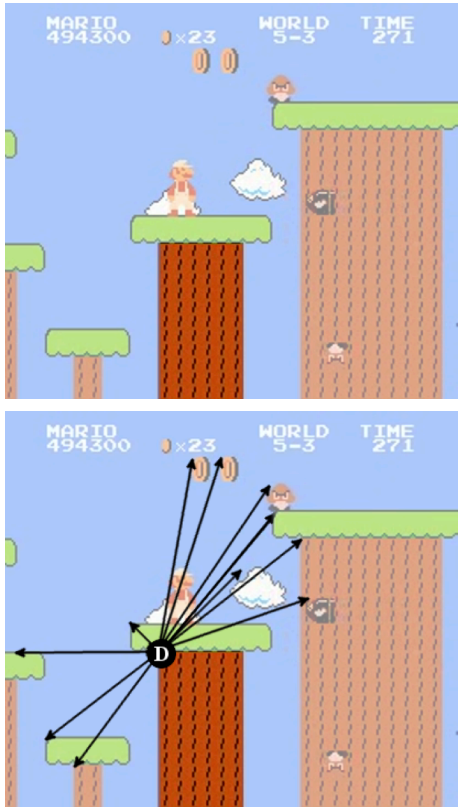

**Figure 3**. A visual representation of a G-Node (top) and its corresponding D-Node (bottom).

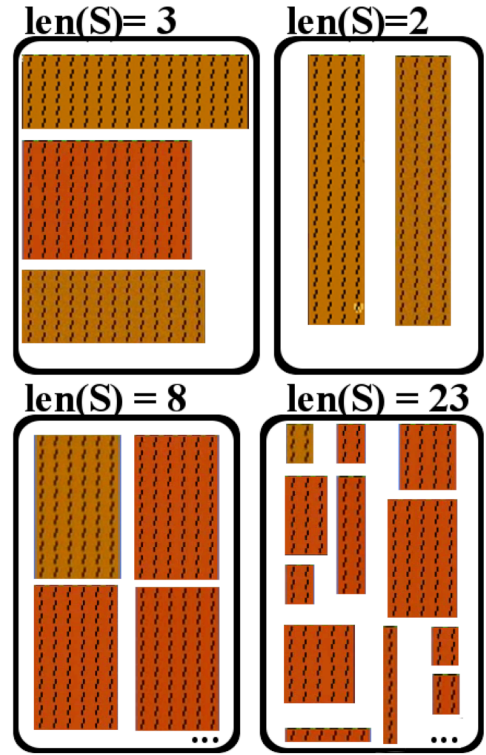

**Figure 4.** A representation of four different S nodes in terms of their shape information.

D nodes contain relational information from a G node to all other G nodes in its level section, regardless of type. This relational information takes the form of an (x, y) vector from the top left corner of each G node, and a second vector to that G nodes center, used in the generation phase. Figure 3 (bottom) represents a D node as the red lines from the G node given in the prior example to all other G nodes in the frame. This node differs the most from the Kalogerakis et al. model; their D node contains discrete information, notably derived from hand-authored edge points between parts of a 3D model. Given our goal to decrease the amount of human authoring required for level section creation, we bypass the human authoring step and instead naively track all possible relationships. In the generation phase we instantiate some of these relationships as edges based on a probabilistic value.

The last of the nodes derived from observable attributes of the initial dataset is the N node. The N node tracks count data of sprites in each frame of our *high interaction* clusters by its sprite type *t* for each originating data point. For example, in Figure 3 our N node stores 251 bark, 31 treetop, 1 goomba, 1 bullet, 1 cloud, and 2 coin sprites. The N nodes' count information serves as a guide during the generation phase. [2]

---

[1] In the original Kalogerakis et al. model, the authors made use of a C node, which contained continuous, mesh information.

[2] Kalogerakis et al. store N values per part, so a humanoid model might have 1, 2, or 4 arms.

The latent variables of the model represent underlying design rules used in the construction of level sections, as derived from the observed values. The S node is a latent variable that represents a category of shapes and their relative placement in a level section. The intuition behind the S node is that it represents a set of interchangeable shapes as learned from observable data. We derive each S node from (G, D) pairs of a sprite type *t*, where the D node tracks all relational information for the given G node. Figure 3 demonstrates one such (G,D) pair.

The S node is derived in two steps. First, the system clusters on (G, D) pairs with the K Means++ clustering approach, with K estimated by the distortion ratio. The distance metric between pairs of (G,D) nodes used by K Means++ is a normalized combination of the different distance metrics for the G and D nodes individually. The distance between two G nodes comes from the edit distance between the two different shapes, which are guaranteed to be of the same type *t*. Between the two D nodes the distance metric is the sum of an iteration across both sets of sorted connections by (x, y) values and the magnitude of the differences between the relational vectors. This gives a sense of the relative position of the paired G node. For example if all the relational vectors point left this would indicate the G node is on the far right of a level section. If one D node has more connections than the other, then our technique simply uses the magnitude of its remaining relational vectors as this is equivalent to the magnitude difference of that vector from a zero vector. This combined distance metric means that even if two G nodes have equivalent shapes, they may end up in different categories if they appear in different relative positions consistently.

In most cases this approach finds that there exists only one S node for each sprite type *t*. For example in the "treetop" category, our approach found a single S node category for bullets, goombas, and koopas used in level sections. However, there exist eight "bark" S nodes. Figure 4 contains a sample of a collection of shape values from four of these S nodes, which contain (g, d) pairs of quantities 2, 2, 3, 3, 3, 8, 9, and 23.

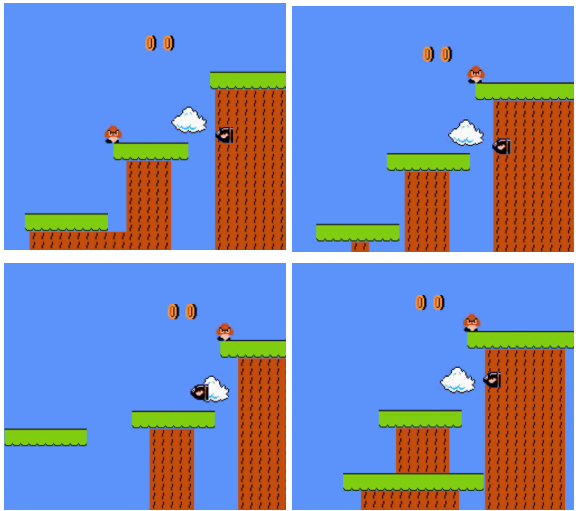

**Figure 5.** Four examples of visualizations of instantiations of the same L node, which used the N node values derived from the level section displayed in Figure 3.

The next step in deriving S nodes is to create a probability table of $P(t_i == edge \mid t_i, d)$, which expresses the probability of some sprite type $t$ being a required edge of another sprite type $t$ given a distance d. By required edge we indicate an equivalent to the Kalogerakis et al. paper for the hand-defined edges that indicate implications between parts. For example, an arm implies a torso. However, since we do not have hand-defined edges we determine these implications probabilistically. We construct the probability table by first collecting all of the vector connections for all D nodes within S. We discretize these connections by their originating shape type $t_i$, their connecting shape type $t_j$, and a discretized value of the vector between them. This discrete value comes from splitting the max possible distance across level sections into 100 buckets. We then determine the likelihood of a given connection's occurrence across all possible connections. We assume this probability to directly relate to the probability that the relationship is an edge, that one shape type at a distance implies another type. Thus our model finds that "leaf" shapes have a high probability to imply bark shapes at low distances.

The top-level latent variable is the L node, which represents a category of level design rules. One can interpret the L node as a template for a specific type of level section design. Figure 5 represents four instantiations of the L node generated from the *high interaction* "treetop" cluster. This node represents a cluster across S nodes and the corresponding N node values from which the values in the S node arise. The clustering at this stage uses K medoids and the Hamming metric across all non-zero entries of an S node's probability table. We make use of K medoids instead of K means at this step to minimize the possibility of combining two distinct clusters of level design rules. This typically leads to only one L node per *high interaction* cluster when there is a clear theme. However, in the case of loosely themed clusters as discussed earlier, this step separates them out. For example, in the case where a *high interaction* cluster ends up with both "lava" level sections and "underwater" level sections due to their joint use of the "wave" sprite, our process recognizes these two as having distinct level design rules and outputs two L nodes and their associated probabilistic models. The earlier clustering of *high interaction* areas based on assumed level design similarities is still important despite this as it vastly decreases the amount of time required to put together the model. Figure 6 demonstrates a representation of the final model this process generates for the "treetop" cluster. The model finds a single N node, a single S node for most sprite types $t$, but multiple S nodes such as those shown in Figure 4.

### 3.3 Generating Level Sections

Generation of a level section is a constraint satisfaction processes wherein combinations of G and D nodes are selected based on requirements and compatibility. The generation algorithm, shown in Figure 7, starts with an empty level section and recursively adds (g, d) pairs until a stopping criterion is met. Recall that G nodes represent a group or shape of sprites of the same type, and D nodes store relative positioning information. The algorithm is seeded with an initial (g, d) pair, and our implementation generates all possible level sections by trying possible (g, d) pairs as starting points. The generation algorithm makes use of two control parameters: $p_E$, which controls the degree to which sections must resemble original sections from video, and $p_C$, which controls how "playable" a level is. These parameters will be described in greater detail below.

When a (g, d) pair is added to a section, the D node indicates a number of potentially required connections to other shapes— roughly the black vectors in Figure 4 (bottom). The algorithm greedily identifies the next best (g, d) pair to add to the section. First, the algorithm determines the type $t$ of the next (g, d) pair to add. There are two possible candidate types. The first candidate type is the next most required sprite type in order to reach the number of sprites of type $t$ as indicated by the closest N node value for the level section. The second candidate sprite type is that that has the highest probability of occurring in the section. The function *get_next_required_edge_type* iterates across each g in

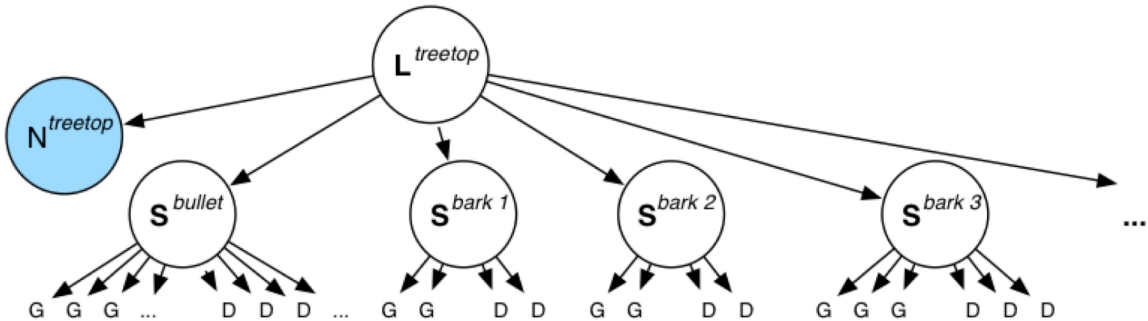

**Figure 6.** A visual representative of the resulting model for this process for the treetop category.

```
void function generate (section, (g, d))
  add_according_to_closest_connection(section, (g, d))
  t_N, d_N = get_next_type_for_nearest_N(section)
  t_E = get_next_required_edge_type(section, p_E)
  next_t = t_N
  if d_N <= 0 and t_E not null
    next_t = t_E
  else if d_N <= 0 and t_E = null
    return section
  for s_next_t in L^t
    for (g, d) in s_next_t
      if coexist_probability(section, (g, d)) > p_C
        generate(section, (g, d))
```

**Figure 7.** The psuedocode for the generation algorithm

the section and uses the probability table from its parent S-node to filter out any possible connections with probability less than $p_E$. From the remaining d node connections, the type with the highest probability is chosen. The value $p_E$ can be understood as a style variance variable, as the lower the value the closer to the original level sections all outputted level sections will be because more relations will be preserved.

Once the type of the sprite is chosen, the algorithm then looks at all S nodes for the type in the model and selects the (g, d) pair that most closely matches the required connections. Matching occurs by looking at the d node to see whether its connections link back to g nodes already placed. The proportion of connections matched by the new (g, d) pair must be above a threshold $p_C$. This value can be understood to be a "playability" variable; the lower the value, the less cohesive and playable the output level section becomes. A playable section is one in which the player can navigate successfully from the left of the screen to the right of the screen according to the mechanics of the game (e.g., there exists no gaps too wide or walls to high to progress).

The algorithm terminates if the current level section meets two conditions. First, if the level section's sprite counts equal or exceed its closest N. Second, if all the required edges as defined by $p_E$ are filled.

The output of the algorithm is a level section represented as a set of (g,d) pairs and the positions of each g node shape. It is converted into a list of sprites and their two-dimensional coordinates by tiling the sprite type across the patterns represented by its g node. This representation is identical to the level section representation used for the original level sections. It is then checked against both the original level sections in the dataset and all output thus far to ensure that it is not a duplicate, using the probabilistic distance measure between frames described in 3.1.

## 4. EVALUATION

We undertook an initial evaluation of our system to measure the effects of varying the variables $p_E$ and $p_C$ on playability and a measure for "Super Mario" style. We make use of the same "treetop" cluster used throughout this paper. We chose to make use of this specific type of level section as it represents a challenge to playability and style. If we made use of any level without the large amount of gaps common in the "treetop" level sections demonstrating playability would have been trivial given that our approach makes use of entire sprite sections from the original (such as an entire ground section). In addition, the "treetop" areas of Super Mario Bros. demonstrate more stylistic variance than the typical level sections in which objects such as pipes are placed on level ground.

We consider a level section *playable* if it meets three conditions: 1) there exists a sprite to the left of the level section that Mario could jump to *from* a platform in a previous section, 2) there exists a sprite to the right of the level section that Mario could jump from *to* in a successive section, 3) there exists a path between these two sprites. Note that we do not consider whether the platforms in the previous or successive section actually exist, but could exist. These conditions were chosen to support integration of generated sections together at a future time. For the final condition we constructed a simple greedy pather. While A* pathing agents cannot complete human-completable Mario levels, we found the agent's behavior sufficiently accurate across a distribution of levels. We vary both $p_E$ and $p_C$ individually (holding the other at 0.1 and 0.8 respectively), sampling twenty level sections from the output and then determining the percentage of the output that is playable. We expect no real difference while varying the "style" variable $p_E$, but significant impact upon playability when varying the "playability" variable $p_C$.

Style is not as easily defined as playability and thus a more difficult feature to measure. We make use of an approximation of style by measuring the distance from a generated section to known section from the gameplay videos as follows. We compute the distance a sprite has to move in a generated level section to match the same sprite type in the closest original level section. This can be understood as a relaxed edit-distance of a sprite for a generated level and its closest originator. We take the average across sprites rather than just summing across the difference for all sprites in order not to favor output levels with fewer total sprites. As with playability we vary both $p_E$ and $p_C$ individually, and sample twenty level sections. We report the median value of this per-sprite edit-distance. A strong correlation between the "style" variable $p_E$ and this value and no correlation between the value and $p_C$ would support our model.

Figure 8 summarizes the results of our initial evaluation of playability. We sampled 20 of the generated level sections from each category to run the tests, given that some variable values lead to many more level sections than others. We report a *percent playable* as the percentage of this sample that were playable according to our playability metric. Figure 8 (top) demonstrates the results of varying the playability variable $p_C$ on our measure of playability. Our model performed as expected, showing a strong correlation between values of the variable and generated level playability (Pearson's r: 0.8778). One interesting and unexpected result can be seen in Figure 8 (bottom) as we vary the "style" variable $p_E$. While the majority of the graph displays no relationship, both values 0.05 and 0.1 resulted in only playable levels. This can be understood as due to the fact that at low values of $p_E$ nearly all possible connections became required, meaning that the results were very similar to the original Super Mario Bros. levels, which are by definition, playable.

Figure 9 summarizes the effects of $p_C$ and $p_E$ on style. Recall that lower values along the y-axis indicated levels closer to the originators given our style metric. Figure 9 (top) shows the result of varying the playability variable resulted in a noisy inverse relationship with the style value (Pearson's r: -0.5831). This can be explained as the style variable was held at 0.1 for these tests, meaning that the space of possible levels was already constrained to output close to the original levels. Further constraining the output to be playable meant that the system generated output that was increasingly similar to the originals, which are also all playable. These results may be due to our explicit check that 90% of the sprites within the frame cannot be within the same position

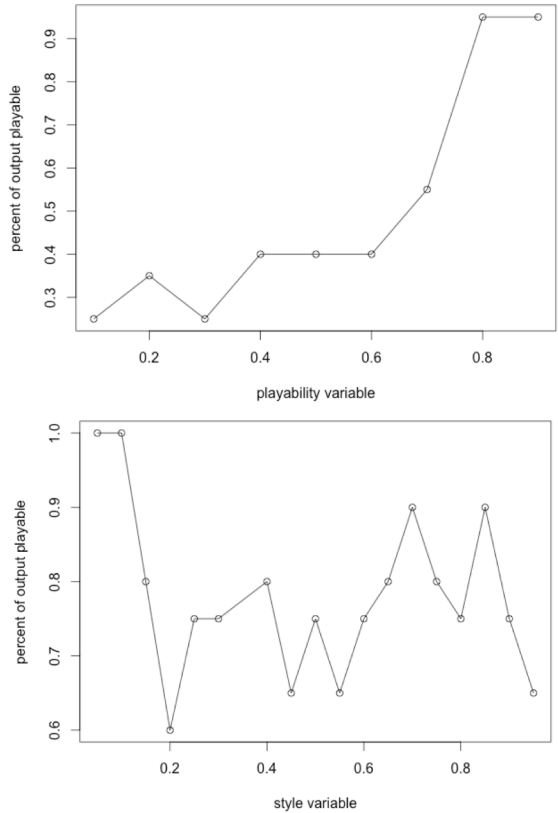

**Figure 8.** The effects of playability (top) and style (bottom) variables on generated level section playability.

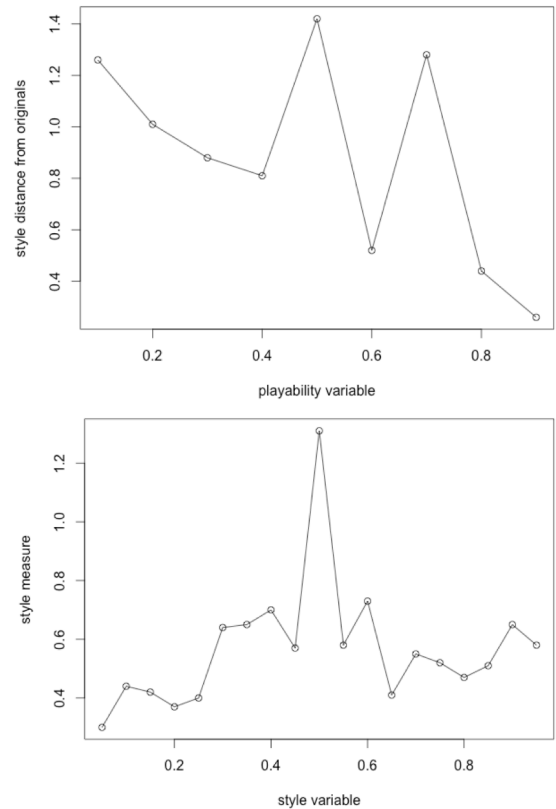

**Figure 9.** The effects of playability (top) and style (bottom) variables on generated level section style.

as mentioned in section 3.1, meaning that we throw out all of the duplicates that would have low edit distances. Figure 9 (bottom) displays no clear relationship between varying the style variable and the median style value. This is a negative result, suggesting that our style metric may be determined by an interplay between variables or that the style variable may have less of an impact than the playability variable

Looking at the output from the system qualitatively we find that while the output with lower values of $p_C$ definitely demonstrates less playability, from our perspective there is more creativity demonstrated in terms of further variation from the original levels' structure. This is a fairly common issue in level generation. We hope to solve this issue in future work by learning game mechanics in addition to game design information from gameplay video, meaning that the system can automatically determine playability of a new section.

In terms of raw counts of output, we saw an increase of close to an order of magnitude with this approach of unique level sections as determined by our difference metric. In particular, we found that with values of $p_C = 0.8$ and $p_E = 0.1$ our system outputted 151 distinct level sections from an original dataset of 17 level sections in the "treetop" category. The output increased as the constraints lessened, with values of $p_C = 0.5$ and $p_E = 0.1$ producing 334 level sections. While we focused on the "treetop" cluster, Figure 10 shows output from six different clusters.

## 5. FUTURE WORK

The clear next step of this process is full level generation. We believe that our process can be extended to learn the orderings of the level section categories it outputs as described in section 3.1. Substituting our own generated level sections into these orderings would allow us to evaluate whether our generated sections cause the same slow down in players seen in the original gameplay videos.

Beyond level generation we seek to learn game mechanics from gameplay video by watching how sprites move or disappear in reference to one another. We have conducted an initial test in a simplified platformer domain using colored blocks that lead to promising results. We will need to extend this process to a real, more complex game like Super Mario Bros. Learning game mechanics is an essential part of learning game design knowledge, and represents a significantly more complex problem than level structure as it involves cause and effect information. Ultimately we hope to extend this process to many different platformers to learn a generalized level design and game mechanics rule-set for the genre. With such a knowledge base we could automatically build entirely new platformers, and extend into further game genres and hybridizations thereof.

## 6. CONCLUSIONS

Procedural content generation algorithms require design knowledge that captures the intuition of human designers about good content. Human authoring of design knowledge remains a large roadblock in procedural content generation and in general video game design for non-experts. We have presented a novel approach to automated learning of computer game design knowledge from gameplay video, with a specific focus on level design knowledge. An initial evaluation of our approach indicates an ability to produce level sections that are both playable and close to the original Super Mario Bros. without hand coding any design criteria. Initial experiments suggest that our approach extends beyond the Super Mario Bros. platformer game and that additional design knowledge such as avatar mechanics may be acquired from gameplay video as well. As gameplay video becomes more accessible and as open machine vision toolkits become more advanced, we see gameplay video as a rich source of design knowledge to be exploited for future procedural content generation and procedural game generation systems.

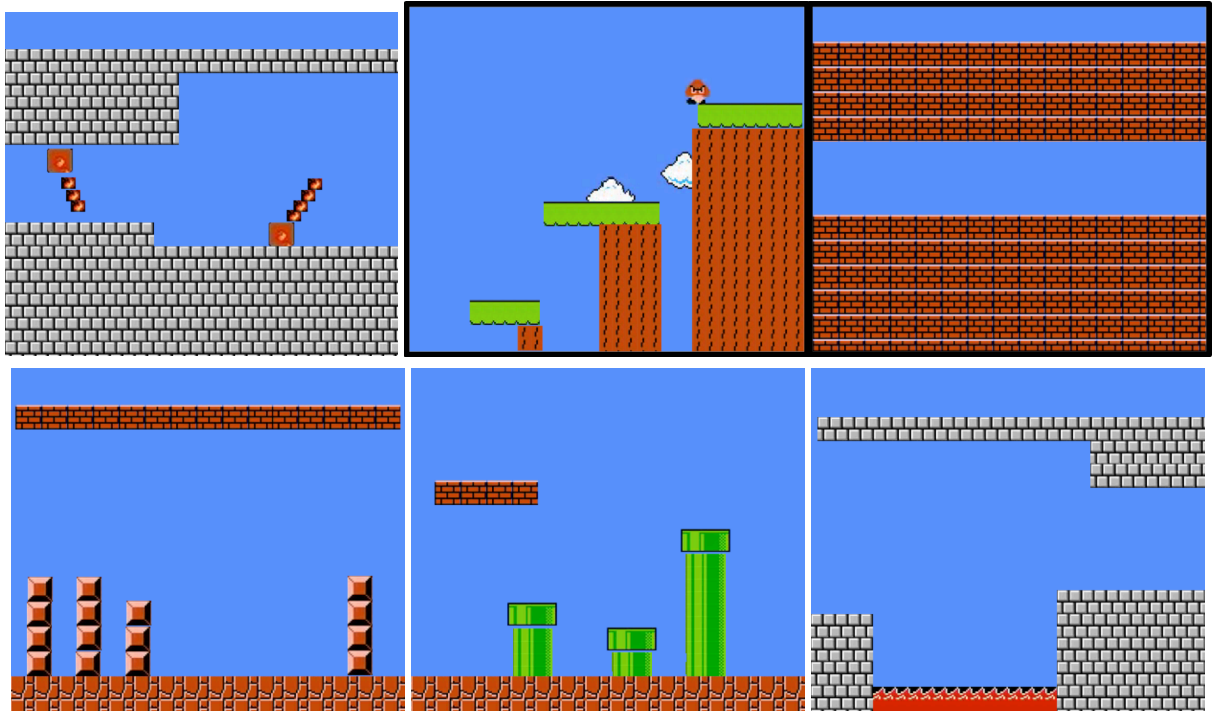

**Figure 10.** Examples of generated level sections from six different *high interaction* categories. Examples outlined in black are failures of the system either stylistically (top middle) or not being a high interaction section (top right).

## 7. ACKNOWLEDGEMENTS

We gratefully acknowledge the NSF for supporting this research under NSF award 1350339.